\documentclass[conference]{IEEEtran}

\usepackage{cite}
\usepackage{amsmath,amssymb,amsfonts}
\usepackage{algorithmic}
\usepackage{graphicx}
\usepackage{textcomp}
\usepackage{xcolor}
\usepackage{amsmath}
\usepackage{amssymb}
\makeatletter
 \let\old@ps@headings\ps@headings
 \let\old@ps@IEEEtitlepagestyle\ps@IEEEtitlepagestyle
 \def\confheader#1{%
 \def\ps@headings{%
 \old@ps@headings%
 \def\@oddhead{\strut\hfill#1\hfill\strut}%
 \def\@evenhead{\strut\hfill#1\hfill\strut}%
 }%
 \def\ps@IEEEtitlepagestyle{%
 \old@ps@IEEEtitlepagestyle%
 \def\@oddhead{\strut\hfill#1\hfill\strut}%
 \def\@evenhead{\strut\hfill#1\hfill\strut}%
 }%
 \ps@headings%
 }
 \makeatother

\confheader{%
 2023 IEEE International Conference on Artificial Intelligence in Engineering and Technology (IICAIET)
 }

 \usepackage[pscoord]{eso-pic}

\newcommand{\placetextbox}[3]{
 \setbox0=\hbox{#3}
 \AddToShipoutPictureFG*{ \put(\LenToUnit{#1\paperwidth},\LenToUnit{#2\paperheight}){\vtop{{\null}\makebox[0pt][c]{#3}}}
 }
 }
  \placetextbox{.14}{0.098}{---------------------}
  \placetextbox{.17}{0.083}{† Corresponding author}
 \placetextbox{.19}{0.069}{* Indicates equal contribution}
 \placetextbox{.23}{0.055}{\small{979-8-3503-0415-2/23/\$31.00~\copyright 2023 IEEE}}

\def\BibTeX{{\rm B\kern-.05em{\sc i\kern-.025em b}\kern-.08em
    T\kern-.1667em\lower.7ex\hbox{E}\kern-.125emX}}
 
\begin{document}

\title{Leveraging Knowledge Graphs for Orphan Entity Allocation in Resume Processing}
\author{\IEEEauthorblockN{
    Aagam Bakliwal \textsuperscript{1}\textsuperscript{*},
    Shubham Manish Gandhi \textsuperscript{1}\textsuperscript{†}\textsuperscript{*},
    Yashodhara Haribhakta \textsuperscript{1}
}
\IEEEauthorblockA{\textsuperscript{1} Department of Computer Science and Engineering\\
COEP Technological University\\
Pune, Maharashtra, India}
}

\maketitle

\begin{abstract}
%Processing and analyzing unstructured data, particularly resumes, pose significant challenges in the field of talent acquisition and recruitment. This research paper presents a novel approach for orphan entity allocation in resume processing using knowledge graphs. Our pipeline integrates many techniques, including association mining, concept extraction, external knowledge linking, named entity recognition, and knowledge graph construction. By leveraging these techniques, our aim is to automate and enhance the efficiency of the job screening process by successfully bucketing orphan entities within resumes. This allows for more effective matching between candidates and job positions, streamlining the resume screening process and enhancing the accuracy of candidate-job matching. Extensive experimentation and evaluation highlight our approach's exceptional effectiveness and resilience, ensuring that alternative measures can be relied upon for seamless processing and orphan entity allocation in case of any component failure. The results of our research highlight the capabilities of knowledge graphs in generating valuable insights through intelligent information extraction and representation specifically in the domain of categorizing orphan entities.

Significant challenges are posed in talent acquisition and recruitment by processing and analyzing unstructured data, particularly resumes. This research presents a novel approach for orphan entity allocation in resume processing using knowledge graphs. Techniques of association mining, concept extraction, external knowledge linking, named entity recognition, and knowledge graph construction are integrated into our pipeline. By leveraging these techniques, the aim is to automate and enhance the efficiency of the job screening process by successfully bucketing orphan entities within resumes. This allows for more effective matching between candidates and job positions, streamlining the resume screening process, and enhancing the accuracy of candidate-job matching. The approach's exceptional effectiveness and resilience are highlighted through extensive experimentation and evaluation, ensuring that alternative measures can be relied upon for seamless processing and orphan entity allocation in case of any component failure. The capabilities of knowledge graphs in generating valuable insights through intelligent information extraction and representation, specifically in the domain of categorizing orphan entities, are highlighted by the results of our research.
\end{abstract}


\section{Introduction}

\subsection{Problem Statement}
%\textcolor{red}{An orphan entity is a piece of information or data within a dataset or context that lacks sufficient surrounding context or clear relationships to other structured data, making it difficult to categorize or determine its specific meaning. In the context of resume processing, an example of an orphan entity could be the term "Python" mentioned in the "Skills" section of a job applicant's resume. Without additional context or explicit information, it is unclear whether "Python" refers to a programming language or the reptile species.}
%Allocating orphan entities, such as skills and qualifications, from a large number of resumes is a time-consuming and error-prone task. The unstructured nature of resumes and the presence of ambiguous references make the allocation of relevant entities challenging for recruiters. Valuable time can be consumed, and the possibility of human errors can be introduced when manually identifying and linking orphan entities. Therefore, an automated and reliable solution is needed to streamline the entity allocation process in talent acquisition and recruitment.

A piece of information or data within a dataset or context that lacks sufficient surrounding context or clear relationships to other structured data is known as an orphan entity, making it difficult to categorize or determine its specific meaning. In the context of resume processing, the term 'Python' mentioned in the 'Skills' section of a job applicant's resume could serve as an example of an orphan entity. Whether 'Python' refers to a programming language or the reptile species remains unclear without additional context or explicit information.

The allocation of orphan entities, such as skills and qualifications, from a large number of resumes, is a time-consuming and error-prone task. The unstructured nature of resumes and the presence of ambiguous references create challenges for recruiters in allocating relevant entities. When manually identifying and linking orphan entities, valuable time can be consumed, and the possibility of human errors can be introduced. Therefore, a need arises for an automated and reliable solution to streamline the entity allocation process in talent acquisition and recruitment.
\subsection{Motivation}
%Motivated by the persistent challenges encountered in orphan entity allocation, we recognize the immense potential of leveraging knowledge graphs as a promising solution. The complex nature of talent acquisition necessitates a structured representation of information, and knowledge graphs offer precisely that. Recognizing the power of knowledge graphs in facilitating entity resolution, semantic linking, and data integration, we aim to explore their application in automating the identification, extraction, and allocation of orphan entities from resumes. Our motivation stems from the desire to enhance the efficiency and accuracy of the talent acquisition process, bridging the gap between valuable but underutilized candidate information and the requirements of employers. By harnessing the capabilities of knowledge graphs, we aim to revolutionize the handling of orphan entities, ultimately streamlining talent acquisition and fostering improved matches between candidates and job opportunities.
The immense potential of leveraging knowledge graphs as a promising solution is recognized, motivated by the persistent challenges encountered in orphan entity allocation. A structured representation of information necessitated by the complex nature of talent acquisition is precisely offered by knowledge graphs. The motivation stems from the desire to enhance the efficiency and accuracy of the talent acquisition process, bridging the gap between valuable but underutilized candidate information and the requirements of employers. Harnessing the capabilities of knowledge graphs, the handling of orphan entities is aimed to be revamped, ultimately streamlining talent acquisition and fostering improved matches between candidates and job opportunities.
\subsection{Objectives}
%We aim to develop a comprehensive pipeline that utilizes knowledge graphs to streamline orphan entity allocation in resume processing. Our approach aims to optimize entity allocation and enhance talent acquisition practices by integrating various techniques such as information extraction and semantic linking. Through concept extraction, association mining, named entity recognition (NER), external knowledge linking, and knowledge graph construction, we aim to improve the precision and reliability of assigning orphan entities in resume processing. This pipeline offers an efficient and reliable solution for employers, facilitating informed decision-making and better alignment between candidate profiles and job requirements.

A comprehensive pipeline, utilizing knowledge graphs to streamline orphan entity allocation in resume processing is aimed to be developed. The precision and reliability of assigning orphan entities in resume processing are aimed to be improved through concept extraction, association mining, named entity recognition (NER), external knowledge linking, and knowledge graph construction. The objective is to establish a streamlined and dependable pipeline that enables well-informed decision-making and enhances the alignment between candidate profiles and job requirements.

\section{Related Work}

Relevant information is extracted from resumes, which consist of structured and unstructured data. The identification and allocation of orphan entities are deemed essential for efficient talent acquisition and recruitment processes.

Three categories encompass traditional approaches for understanding orphan entities: sequence labeling-based, hypergraph-based, and span-based methods.

The sequence labeling approach \cite{trad1} is used to predict labels for each token, but it struggles with nested NER. Some works such as \cite{trad2}\cite{trad3} have adapted the sequence labeling model for nested entity structures by designing a special tagging scheme. Span-based methods first extract spans through enumeration \cite{span1} or boundary identification \cite{span2}, and then classify the spans. For example, \cite{span3} proposes a two-stage identifier that first locates entities and then labels them, treating NER as a joint task of boundary regression and span classification. However, these methods mainly focus on boundary identification, which might not be necessary for short orphan entities.

A standard method for inferring word meanings by matching predefined keywords or dictionaries is keyword matching. While providing some level of disambiguation, it may have limited accuracy and fail to capture complete semantic context \cite{phdthesis}.

In the context of resume information extraction, a two-step algorithm for extracting information from resumes was proposed by \cite{hanne2018two}. The Writing Style feature is used to first identify resume blocks, and then factual information attributes are identified. Though promising results were obtained on real-world datasets, it falls short in extracting word meanings beyond section names. The effectiveness of cascaded processing and multiple passes through resumes was demonstrated by \cite{yu-etal-2005-resume}, outperforming the multi-pass flat model approach. A framework using text classifiers to build a structured resume repository, facilitating knowledge extraction and efficient data organization, was presented by another study \cite{10.1007/978-3-319-21042-1_58}.

To enhance orphan entity allocation, the use of ontologies \cite{10.1007/978-3-319-21042-1_58} was explored. These systems leveraged ontology-defined hierarchies, concepts, and relationships to improve the extraction of contextual information from resumes. Furthermore, orphan entities are extracted through the combination of word embeddings and entity co-occurrence analysis by \cite{Zu2019ResumeIE}, capturing semantic similarity and contextual information and relationships within the resume.

Knowledge Graph-Augmented Abstractive Summarization with Semantic-Driven Cloze Reward \cite{DBLP:journals/corr/abs-2005-01159} utilizes semantic knowledge from knowledge graphs to enhance entity allocation accuracy, however, it is resource-intensive and has a complex implementation. 

A permutation-based bidirectional training approach is employed by \cite{XLNet} for the identification of named entities. This approach is incorporated into our NER module, alongside the foundational idea from \cite{hanne2018two}, where a customized cascading model is introduced, specifically designed for resumes. The existing knowledge generated by \cite{speer2018conceptnet} is leveraged, combined with Association Mining techniques, to tailor our algorithm for optimal processing of resumes.

\section{Proposed Pipeline}

A comprehensive pipeline for identifying orphan entities in resumes is presented, addressing the limitations of pre-trained language models like the Bidirectional Encoder Representations from Transformers (BERT), which capture general language representations but lack domain-specific knowledge \cite{Liu_Zhou_Zhao_Wang_Ju_Deng_Wang_2020}. The pipeline involves a combination of modules, namely, concept mining, association mining, NER, external knowledge linking, and knowledge graph creation.

For each module in the pipeline, an orphan entity and the corresponding resume are taken as input, and an output is generated. Through thorough investigation and experimentation using various thresholds for measuring the similarity between the output and the orphan, a predefined minimum threshold has been established for each module. If the distance between the orphan entity and the identified output exceeds this threshold, the output is incorporated into the knowledge graph. Otherwise, the next module is executed.

The pipeline begins with a concept extraction module, where the orphan and the corresponding resume are analyzed to extract relevant concepts. If the distance between the concept and the orphan is below the predefined threshold, the identified entity is added to the knowledge graph.

If satisfactory results are not achieved by the concept extraction module, association mining is conducted using Apriori analysis and word similarity. This process discovers associations between the orphan entity and words present in resumes and identifies the most similar word to input to the knowledge graph. If the similarity of the identified word exceeds the predefined threshold for that module, that word is added to the knowledge graph.

If both the concept extraction and association mining modules fail to yield satisfactory results, NER is employed. NER identifies potential sources in the resume and finds the most similar word from the knowledge graph. If the similarity surpasses the predefined threshold, the closest word (source entity) is added to the knowledge graph.

Finally, even if an output is produced by the NER module whose distance is less than the orphan for the predefined threshold, the output of the external knowledge linking module is searched to check whether the orphan exists in it.

Throughout execution, the knowledge graph is updated with new entities and edges whenever a suitable source entity is identified.
\subsection{Concept Mining}
%The concept extraction module in this research paper focuses on extracting meaningful concepts from artifacts. Inspired by an innovative approach to concept mining \cite{8594851}, the module utilizes embedding vector representations to analyze occurrence contexts, evaluate concept quality, and assess their suitability within local contexts.

%To initiate the module, Glove vectors\cite{glove}, pre-trained word embeddings, are initialized. These vectors enable the module to find similar or associated words, enhancing concept mining accuracy. The module utilizes a framework created by \cite{speer2018conceptnet} to extract relevant related entities associated with orphan entities.

%For concept extraction, the module takes orphan entities as input and utilizes the ConceptNet platform \cite{speer2018conceptnet} to generate a set of potentially related words. By calculating the distance between context words extracted from artifacts and the related words using Glove vectors, the module selects the concept with the minimum distance below the pre-defined threshold. The module outputs a tuple, pairing the orphan entity with its corresponding concept.

%In case the concept extraction module does not produce satisfactory outcomes, the subsequent module, namely Association Mining, is executed.

%The workflow of this module is illustrated in Figure \ref{fig:concept}.

The concept extraction module extracts meaningful concepts from artifacts and utilizes embedding vector representations to analyze occurrence contexts, evaluate concept quality, and assess their suitability within local contexts, inspired by an innovative approach to concept mining \cite{8594851}.

Glove vectors\cite{glove}, which are pre-trained word embeddings, are initialized. With the help of these vectors, similar or associated words can be found, thereby enhancing accuracy. A framework created by \cite{speer2018conceptnet} is implemented to extract relevant related entities associated with orphan entities.

The Resume along with the orphan entities are taken as input, and the ConceptNet platform \cite{speer2018conceptnet} is utilized to generate a set of potentially related words. The distance between context words extracted from artifacts and the related words is measured using Glove vectors, and the concept with the minimum distance below the pre-defined threshold is selected by the module. A tuple is returned by the module, pairing the orphan entity with its corresponding concept.

If the concept extraction module does not produce satisfactory outcomes, the subsequent module, namely Association Mining, is executed.

\begin{figure}  
  \centering
  \includegraphics[width=\columnwidth]{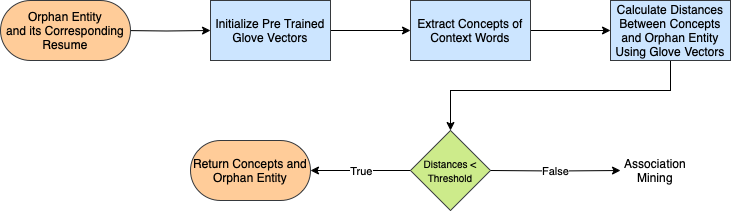}
  \caption{Concept Mining Module Workflow}
  \label{fig:concept}
\end{figure}

The workflow of this module is illustrated in Figure \ref{fig:concept}.

\subsection{Association Mining}
\begin{figure*}[ht]
  \centering
  \includegraphics[width=\textwidth]{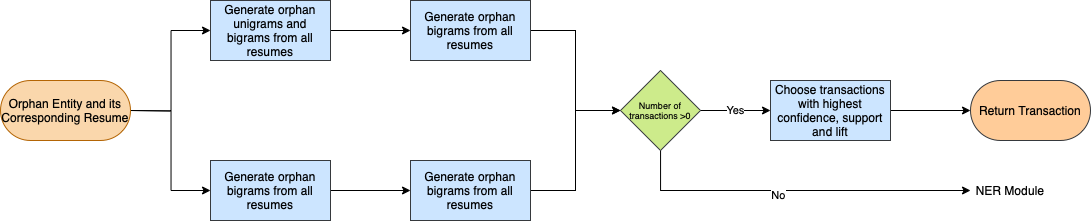}
  \caption{Association Mining Module Workflow}
  \label{fig:association_mining}
\end{figure*}
%Our approach in this study draws inspiration from the challenges addressed by the Apriori algorithm, as demonstrated in the study focused on e-commerce applications and the extraction of association rules from customer purchase transactions \cite{riyadi2019analysis}. We adapt the Apriori algorithm to uncover word associations in our specific context of interest.

%For the implementation of this approach, a module has been devised utilizing the Apriori algorithm. The module initially extracts contextual words pertaining to the orphan from the resume and generate transaction lists for uni-grams and bi-grams, where the transaction lists are created from the context words and encompass the entire corpus of resumes within the dataset. Subsequently, the module employs the Apriori algorithm to process these transaction lists, filtering the results based on the support, confidence, and lift, as defined in \cite{definitions}. 

 %In case the distances between the results produced by the module and the orhpan entity are above a specified threshold, the subsequent module, namely NER, is executed, else the results are added to the knowledge graph.

%The workflow of this module is illustrated in Figure \ref{fig:association_mining}.

Inspiration is drawn from the challenges addressed by the Apriori algorithm in the study focused on e-commerce applications and the extraction of association rules from customer purchase transactions \cite{riyadi2019analysis}. In our specific context of interest, we adapt the Apriori algorithm to uncover word associations.

A module has been devised that utilizes the Apriori algorithm. Initially, contextual words pertaining to the orphan are extracted from the resume, and transaction lists for uni-grams and bi-grams are generated. These transaction lists are created from the context words and encompass the entire corpus of resumes within the dataset. Subsequently, the Apriori algorithm is employed to process these transaction lists, and the results are filtered based on the support, confidence, and lift, as defined in \cite{definitions}.

If the distances between the results produced by the module and the orphan entity are above a specified threshold, the subsequent module, namely NER, is executed. Otherwise, the results are added to the knowledge graph.

The workflow of this module is illustrated in Figure \ref{fig:association_mining}.
\subsection{Named Entity Recognition}

\begin{figure}[h!]
  \centering
  \includegraphics[width=\columnwidth]{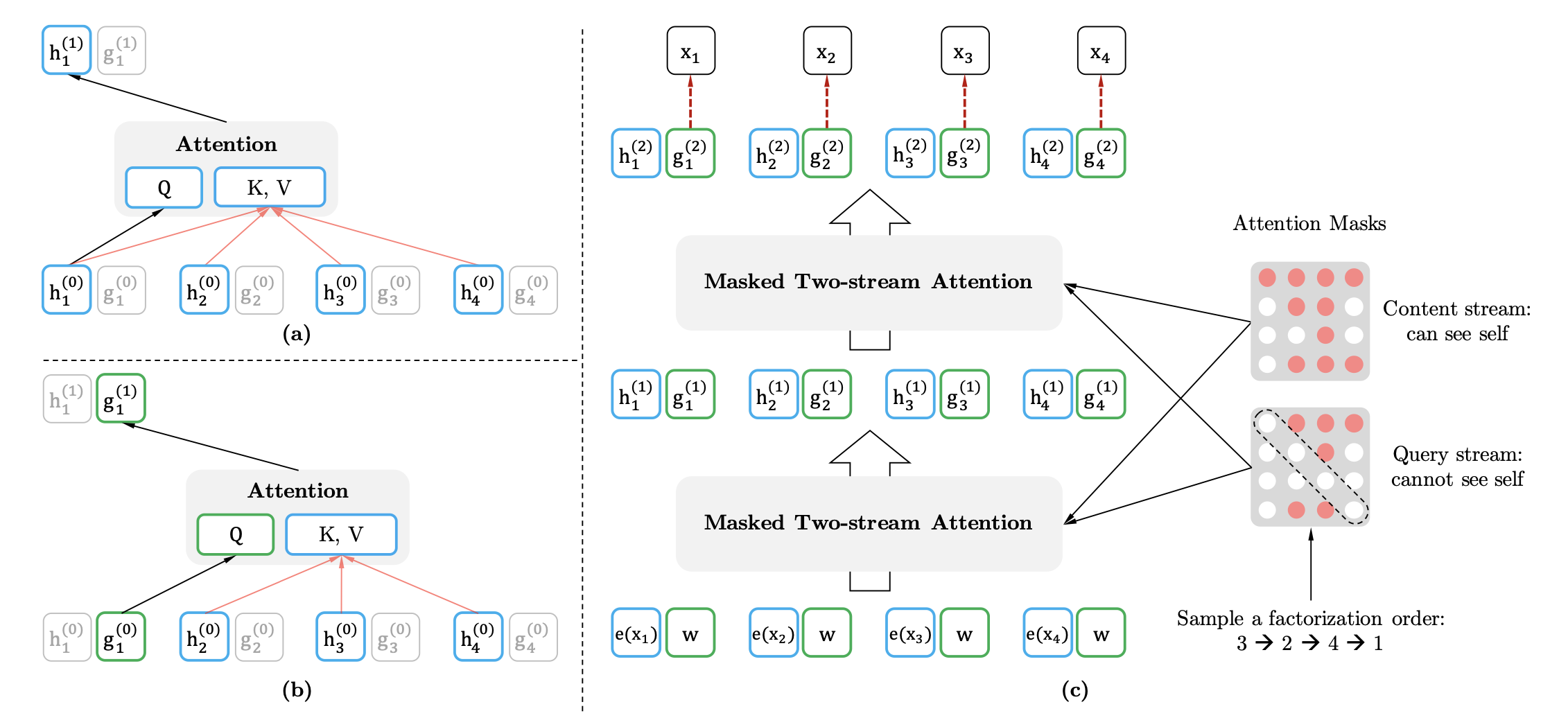}
  \caption{XLNet model architecture as proposed in \cite{XLNet}}
   \label{fig:NER}
\end{figure}

A pivotal role is assumed by the NER module, leveraging the XLNet model \cite{XLNet} to facilitate the identification and classification of named entities within textual data . By providing XLNet with a target word for identification and its encompassing context words, linguistic nuances are grasped in a dynamic, bidirectional manner. XLNet distinguishes itself from other models by implementing the "permutation-based" training approach \cite{XLNet}. This unique approach allows the limitations of previous models, such as BERT and roBERTa, to be overcome, with better capture of long-range dependencies and bidirectional context \cite{XLNet}. Given that the contextual words in a resume hold extreme importance in our task, the use of XLNet is proposed by the method. This model holds the promise of cultivating a deeper comprehension of the input word within its expansive linguistic context, ultimately leading to heightened accuracy and robustness in NER results.

The workflow of this module is illustrated in Figure \ref{fig:NER}.
\color{black}

\subsection{External Knowledge Linking} 

%To enhance the understanding and context of the orphan entities, our pipeline incorporates external knowledge linking, which establishes connections between entities, concepts, or terms in the dataset and relevant external knowledge bases. In this specific case, we utilize external knowledge from Coursera and LinkedIn. This approach builds upon previous techniques that leverage external knowledge to enhance entity comprehension and connectivity in datasets using Wikipedia \cite{torisawa2007exploiting}.

%By integrating external knowledge linking, we enrich our knowledge graph with valuable information, particularly regarding the relationships between entities. This step is not a one-time occurrence but rather a module that runs regularly to ensure the continuous update of our Knowledge Graph with emerging and up-to-date skills.

%Through the utilization of external knowledge linking, we extend the breadth and depth of our research. By tapping into external knowledge sources, we gain access to a wealth of additional insights and information. This iterative process enables us to keep our Knowledge Graph current and relevant, ensuring that it remains a comprehensive and dynamic resource for understanding the relationships and connections within our dataset.

%The workflow of this module is illustrated in Figure \ref{fig:ext_knowledge_linking}.

External knowledge linking is incorporated to enhance the understanding and context of the orphan entities. This process establishes connections between entities, concepts, or terms in the dataset and relevant external knowledge bases, specifically from Coursera and LinkedIn. This approach builds upon previous techniques that leverage external knowledge, such as Wikipedia, to enhance entity comprehension and connectivity in datasets\cite{torisawa2007exploiting}.

By integrating external knowledge linking, we enrich our knowledge graph with valuable information, particularly regarding the relationships between entities. This step is not a one-time occurrence but rather a module that runs regularly to ensure the continuous update of our Knowledge Graph with emerging and up-to-date skills.

Moreover, we gain access to a wealth of additional insights and information through various external knowledge sources and the iterative nature enables us to keep our Knowledge Graph current and relevant, ensuring that it remains a comprehensive and dynamic resource for understanding the relationships and connections within our dataset.

\begin{figure} [h!] 
  \centering
  \includegraphics[width=\columnwidth]{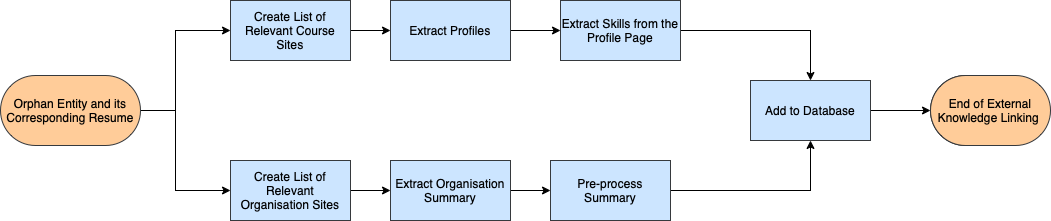}
  \caption{External Knowledge Linking Module Workflow}
  \label{fig:ext_knowledge_linking}
\end{figure}

The workflow of this module is illustrated in Figure \ref{fig:ext_knowledge_linking}.

\subsection{Knowledge Graph Building}
%In our research, knowledge graphs serve as the fundamental backbone for our orphan entity allocation system. By utilizing a source-destination tuple, we seamlessly integrate newly discovered relationships into the knowledge graph through either Concept Mining, Association Mining, NER or external knowledge linking modules. This approach allows us to visualize the connections between entities, facilitating rapid searches and accurate identification of relationships. The knowledge graph acts as a comprehensive representation of the underlying data, empowering us to efficiently navigate and explore the intricate network of relationships within the system efficiently. 

%The power of knowledge graphs becomes evident as they enable the efficient and rapid allocation of orphans which are already present in the graph. By leveraging the existing knowledge graph, we determine the context words of the orphan from the resume and calculate the minimum cosine distance between their glove vectors and the neighboring words in the knowledge graph. This approach streamlines the allocation process, as the hard work of building the knowledge graph beforehand simplifies the task, which forms the cornerstone of our novel methodology.

%A subset of the knowledge graph is illustrated in Figure \ref{fig:kgdiagram}.
Knowledge graphs serve as the backbone for our proposed methedology. A source-destination tuple seamlessly integrates newly discovered relationships into the knowledge graph through modules like Concept Mining, Association Mining, NER, or External Knowledge Linking. This approach allows us to visualize the connections between entities, facilitating rapid searches and accurate identification of relationships. The knowledge graph acts as a structured comprehensive representation of the underlying data, empowering us to efficiently navigate and explore the intricate network of relationships within the system.

The power of knowledge graphs becomes evident as they enable the efficient and rapid allocation of orphans that are already present in the graph. By leveraging the existing knowledge graph, we determine the context words of the orphan from the resume and return the word with the minimum cosine distance between their glove vectors and the neighboring words in the knowledge graph.

\begin{figure}[h!]
  \centering
  \includegraphics[width=\columnwidth]{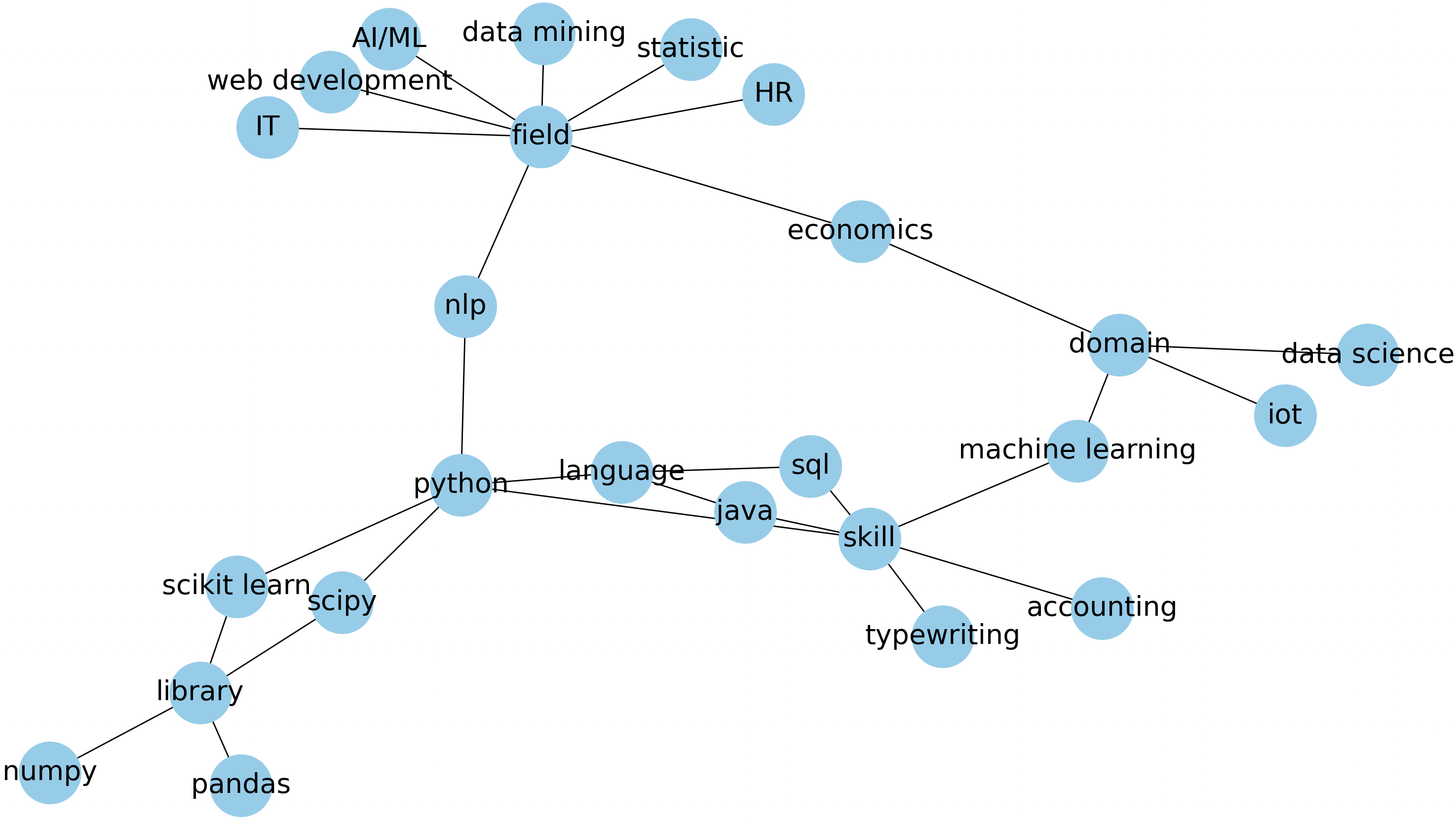}
  \caption{Subset Graph within the Knowledge Graph}
   \label{fig:kgdiagram}
\end{figure}

A subset of the knowledge graph is illustrated in Figure \ref{fig:kgdiagram}.

\section{Experimental Setup}
\subsection{Dataset Description}
%The dataset utilized in this paper\cite{dutt_resume_dataset}, encompasses a comprehensive assortment of resumes, featuring a substantial sample size exceeding 900. These resumes span a wide array of professional backgrounds, including testing, web design, HR (human resources), and various others. The diverse nature of this dataset, originating from different domains, facilitates a comprehensive examination and assessment of the proposed techniques for mapping orphan entities using knowledge graphs within the realm of resume processing. With its extensive collection of resumes, this dataset presents a robust and representative compilation of real-world instances, capturing the intricacies and variabilities associated with distinct job roles and industries. Consequently, it serves as an excellent foundation for the evaluation of the proposed methods for mapping orphan entities using knowledge graphs in the context of resume processing.
The dataset utilized in this paper \cite{dutt_resume_dataset} encompasses a comprehensive assortment of resumes, with a substantial sample size exceeding 900 encompassing a wide array of professional backgrounds, including but not limited to testing, web design, HR (human resources). The diverse nature of this dataset, originating from different domains, facilitates a comprehensive assessment of the proposed pipeline. With its extensive collection of resumes, a robust and representative compilation of real-world instances is presented, capturing the intricacies and variabilities associated with distinct job roles and industries.

\subsection{Experimental Methodology}
%\paragraph{Dataset Preparation}The dataset preparation process involved several steps to preprocess the resumes for analysis. Firstly, all resumes were transformed to lowercase for consistency and to avoid discrepancies in capitalization. 

%Next, the resumes were tokenized, which involved splitting the text into individual words or tokens. Tokenization allowed for a more detailed analysis of the textual data by breaking it down into its constituent parts.

%Stopwords, common words like "the" and "and" that carry little meaning, were removed to enhance the quality and relevance of the textual data. By eliminating stopwords, the focus shifted to more informative words and phrases in the resumes.

%To further refine the textual data, stemming and lemmatization techniques were applied. Stemming involved removing word suffixes to obtain their root form, while lemmatization transformed words into their dictionary or lemma form. These techniques helped consolidate word variations, reducing redundancy and improving the accuracy of the analysis by treating related words as the same.

%The combination of lowercase transformation, tokenization, stopwords removal, stemming, and lemmatization ensured that the textual data from the resumes were appropriately processed and prepared for subsequent analysis. The Text-Processing workflow is illustrated in Figure \ref{fig:textpre}.

\paragraph{Dataset Preparation}The dataset preparation process involves several steps to preprocess the resumes for analysis. Firstly, the lowercase transformation was applied to all resumes for consistency and to avoid discrepancies in capitalization.

Next, the text of the resumes was tokenized, which involved splitting it into individual words or tokens. Tokenization allowed for a more detailed analysis of the textual data by breaking it down into its constituent parts.

To enhance the quality and relevance of the textual data, stopwords, common words like "the" and "and" that carry little meaning, were removed. By eliminating stopwords, the focus shifted to more informative words and phrases in the resumes.

To further refine the textual data, stemming and lemmatization techniques were employed. Word suffixes were removed to obtain their root form through stemming, while lemmatization transformed words into their dictionary or lemma form. These techniques helped consolidate word variations, reducing redundancy, and improving the accuracy of the analysis by treating related words as the same.

\begin{figure}[h!]
  \centering
  \includegraphics[width=\columnwidth]{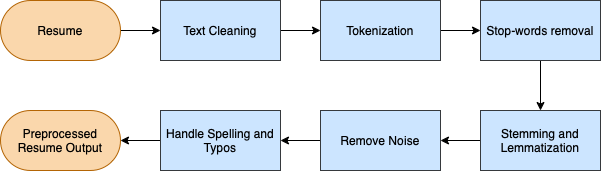}
  \caption{Text-preprocessing Workflow}
  \label{fig:textpre}
\end{figure}

The Text-Processing workflow is illustrated in Figure \ref{fig:textpre}.

% In addition to the pre-processing steps mentioned in \ref{fig:textpre}, noise removal techniques were also applied to further refine the textual data. Short words with minimal contribution to the overall meaning or context of the resumes were filtered out. This noise removal step helped reduce irrelevant or insignificant information, allowing for a focus on more substantial and meaningful content.

\paragraph{Knowledge Graph Building} 

%Initially, the orphan entity is presented to the first module, concept mining. If the module fails to return a relevant entity (based on the predefined threshold), the system proceeds to the next module in the pipeline.h Upon encountering a module that successfully returns a relevant node, the orphan entity is then attached to this node in the knowledge graph.
The first module, concept mining, is where the orphan entity is initially presented along with the resume. If a relevant entity is not returned by the module (based on the predefined threshold), the system proceeds to the next module in the pipeline. The orphan entity is then attached to this node in the knowledge graph upon encountering a module that successfully returns a relevant node.

\begin{figure}[h!]
  \centering
  \includegraphics[width=\columnwidth]{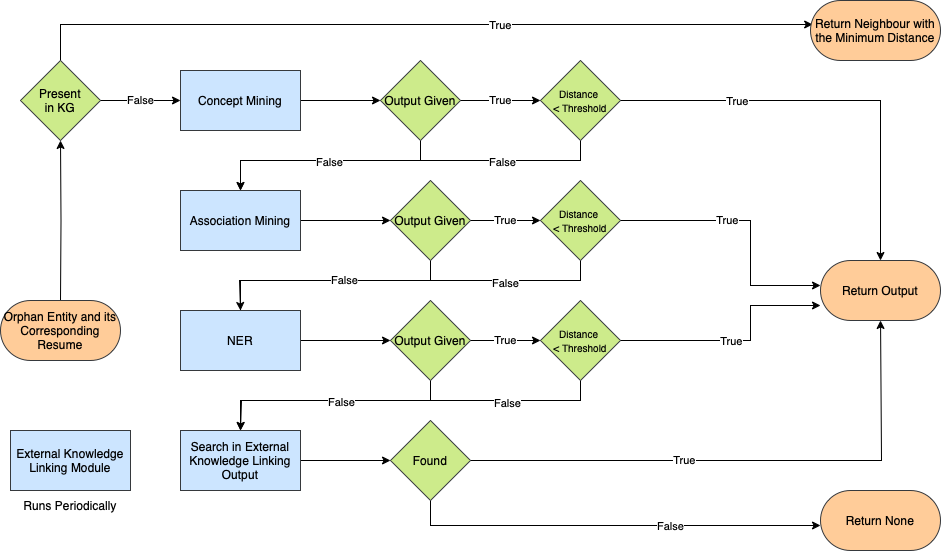}
  \caption{Overall Program Workflow}
   \label{fig:maindiagram}
\end{figure}

\paragraph{Integration} %The proposed approach for our research consists of an integration module that incorporates four distinct models: Concept Mining, Association Mining, NER, and External Knowledge Linking.
% These models have been synergistically merged within a hierarchical pipeline to harness their individual capabilities collectively. This hierarchical model execution as illustrated in figure \ref{fig:maindiagram}  allows for the seamless flow of information and knowledge extraction throughout the various stages of the analysis process.
Incorporating four distinct modules: Concept Mining, Association Mining, NER, and External Knowledge Linking, these modules have been synergistically merged within a hierarchical pipeline to harness their individual capabilities collectively. The hierarchical model execution, as illustrated in Figure \ref{fig:maindiagram}, allows for the seamless flow of information and knowledge extraction throughout the various stages of the analysis process.

\section{Results and Analysis}

\subsection{Algorithm Accuracy}

%After successive training on the dataset provided in \cite{dutt_resume_dataset}, our research endeavors resulted in an algorithm that demonstrated remarkable precision in categorizing orphan entities. The corresponding accuracies are outlined in Table \ref{table:accuracy}, computed using the equation (\ref{eq:accuracy}). Notably, due to the use of knowledge graphs, there was a consistent improvement in the algorithm's accuracy as the training dataset expanded and the number of training epochs increased. These findings emphasize the potential for further refining and optimizing the algorithm's performance by incorporating supplementary resume data.

%In this research paper, we utilize a specific formula to measure the accuracy of orphan entity allocation. The formula expressed mathematically, is as follows:
After successive training on the dataset provided in \cite{dutt_resume_dataset}, an algorithm was developed and remarkable precision was displayed in categorizing orphan entities. The corresponding accuracies are outlined in Table \ref{table:accuracy}, computed using the equation (\ref{eq:accuracy}). Notably, due to the use of knowledge graphs, there was a consistent improvement in the algorithm's accuracy as the training dataset and  training epochs expanded. These findings emphasize the potential for further refining and optimizing the algorithm's performance by incorporating supplementary resume data.

A specific formula is utilized to measure the accuracy of our methodology. The formula expressed mathematically is as follows:
\begin{equation}
\textit{accuracy} = \frac{\textit{Number of orphans correctly allocated}}{\textit{Total number of orphans allocated}} \times 100 \label{eq:accuracy}
\end{equation}
where the accuracy is computed by dividing the number of correctly allocated orphan entities by the total number of orphan entities.

\begin{table}[htbp]
  \centering
  \caption{Accuracy on different Datasets}
  \label{table:accuracy}
  \begin{tabular}{|c|c|}
    \hline
    Dataset & Accuracy \\
    \hline
    Resume & 86.32\% \\
    Non-resume specific text corpus & 78.85\% \\
    \hline
  \end{tabular}
\end{table}

\subsection{Noteworthy Observations}
\subsubsection{Complementary Effect of Concept Mining and Association Mining}

%Our investigation revealed an interesting interplay between the concept mining and association mining algorithms. When incorporating each resume, we observed that if the concept mining algorithm failed to yield significant results, the association mining algorithm became progressively more effective. This suggests a complementary relationship between the two approaches, where the limitations of one method can be mitigated by the strengths of the other. By leveraging this interdependence, our research demonstrates the potential for enhanced accuracy and effectiveness in orphan entity allocation.
The potential for enhanced accuracy and effectiveness in orphan entity allocation is demonstrated. Showcasing the complementary relationship between concept mining and association mining. When each resume is incorporated, it was observed that if significant results were not yielded by the concept mining algorithm, the association mining algorithm became progressively more effective. This suggests that the strengths of one method can mitigate the limitations of the other, highlighting the potential for leveraging this interdependence to achieve improved orphan entity allocation.
\subsubsection{Influence of Algorithm Sequence}

%Another noteworthy observation pertains to the sequencing of the concept mining, association mining, and NER algorithms. Specifically, when the concept mining and association mining algorithms did not yield results within a specified threshold distance, the NER algorithm consistently achieved high accuracy in predicting the type of entity. However, when the NER module was executed prior to the association mining algorithm, the results generated by the NER algorithm were seldom correct. This finding emphasizes the importance of executing the algorithms in a specific order to ensure optimal performance and accurate allocation of named entities in resumes.
A noteworthy observation pertains to the sequencing of concept mining, association mining, and NER. Specifically, when results were not yielded within a specified threshold distance by the concept mining and association mining algorithms, high accuracy in predicting the type of entity was consistently achieved by NER. However, when the NER module was executed prior to the association mining algorithm, the results generated by the NER algorithm were seldom correct. This finding emphasizes the importance of executing the algorithms in a specific order to ensure optimal performance and accurate allocation.
\subsection{Contributions of the Research} 

%The implications of our findings suggest that by utilizing structured representations and semantic connections, knowledge graph-based approaches have the potential to greatly enhance the accuracy and efficiency of orphan entity allocation. This, in turn, can streamline the talent acquisition process, facilitating better matches between candidates and job opportunities.

%This research has made significant contributions to the advancement of knowledge-driven applications in talent acquisition and recruitment. The key contributions of our research can be summarized as follows:
The implications of our findings suggest that by utilizing structured representations and semantic connections, the potential for greatly enhancing the accuracy and efficiency of orphan entity allocation is demonstrated by knowledge graph-based approaches. This, in turn, can lead to the streamlining of the talent acquisition process, facilitating better matches between candidates and job opportunities.

Significant contributions to the advancement of knowledge-driven applications in talent acquisition and recruitment have been made by this research. The key contributions can be summarized as follows:

%\begin{enumerate}
 %   \item Integrated Solution: Our developed solution takes an integrated approach to comprehensively address the challenge of orphan entity allocation in resume processing. It utilizes four distinct modules that synergistically tackle each other's weaknesses, resulting in a robust and effective solution for automating the allocation of orphan entities.
  %  \item Automation and Efficiency: Our research demonstrates the effectiveness and efficiency of knowledge graph-based approaches in automating orphan entity allocation. By leveraging the power of knowledge graphs, we are able to capture complex relationships and produce accurate results quickly and efficiently.
  %  \item Optimization of Recruitment Process: Our research contributes to optimizing the recruitment process through intelligent information extraction and representation. By automating orphan entity allocation and improving the accuracy of entity mapping, our pipeline enables better matches between candidates and job requirements.
%\end{enumerate}
\begin{enumerate}
\item An integrated approach is taken by our developed solution to comprehensively address the challenge of orphan entity allocation in resume processing. Four distinct modules are utilized, which synergistically tackle each other's weaknesses, resulting in a robust and effective solution for automating the allocation of orphan entities.
\item The effectiveness and efficiency of knowledge graph-based approaches in automating orphan entity allocation are demonstrated by our research. By leveraging the power of knowledge graphs, complex relationships are captured, and accurate results are produced quickly and efficiently.
\item An optimization of the recruitment process is contributed by our research through intelligent information extraction and representation. By automating orphan entity allocation and improving the accuracy of entity mapping, better matches between candidates and job requirements are enabled by our pipeline.
\end{enumerate}
Overall, our research provides valuable insights into the application of knowledge graphs in talent acquisition and recruitment. By bridging the gap between candidate information and job requirements, our approach enhances the recruitment process and contributes to the advancement of knowledge-driven applications in this domain.

\subsection{Limitations and Future Research Directions} %Our research on orphan entity allocation in resume processing presents a promising approach. However, certain limitations warrant further investigation. The quality and completeness of the knowledge graph and external knowledge sources greatly influence the effectiveness of our pipeline. Future research should prioritize improving the accuracy of our pipeline using a limited dataset. While our pipeline effectively addresses orphan entity allocation in resume processing, it may also yield excellent results in other domains, warranting further exploration. Additionally, incorporating machine learning and deep learning techniques can enhance entity classification, disambiguation, and linking accuracy. Future research should focus on integrating advanced Artificial Intelligence (AI) models to further enhance the performance and effectiveness of the pipeline.
A promising approach is presented for orphan entity allocation in resume processing. However, certain limitations warrant further investigation. The effectiveness of our pipeline is greatly influenced by the quality and completeness of the knowledge graph and external knowledge sources. Improving the accuracy of our pipeline using a limited dataset should be prioritized in future research. Furthermore, excellent results in other domains may be yielded by our pipeline, warranting further exploration. Additionally, entity classification, disambiguation, and linking accuracy can be enhanced by incorporating machine learning and deep learning techniques. Future research should focus on integrating advanced Artificial Intelligence (AI) models to further enhance the performance and effectiveness of the pipeline.

\section{Conclusion}

In conclusion, orphan entity allocation in resumes was emphasized in our research using a combination of concept mining, association mining, and named entity recognition algorithms. Through experiments and analysis, several significant findings that contribute to the understanding and improvement of this domain have been made.

Firstly, promising accuracy was demonstrated by our algorithm, achieving approximately 86\% when trained on a dataset. This indicates the potential for effectively allocating named entities in resumes using our approach. Moreover, the accuracy of the algorithm consistently improved with the inclusion of more resumes and iterations, highlighting the value of continuous training and refinement.

A complementary connection between concept mining and association mining algorithms was revealed by our research, where the weaknesses of one approach were counterbalanced by the strengths of the other. Moreover, the crucial role of algorithm sequencing was highlighted by our study, underscoring the significance of meticulously deliberating the execution order to enhance the allocation process.

It is important to note that our algorithm is optimized for resume text and does not achieve the same level of accuracy when applied to general text corpora. Therefore, customizing the approach to align with the unique characteristics and linguistic nuances of different text domains becomes essential.

Overall, valuable insights and methodologies for the effective allocation of orphan entities in resumes are provided by our research. The findings can be built upon in future studies to further enhance the accuracy and applicability of such algorithms in the field of resume analysis and information retrieval.

% \subsection{Future Work}
% Further research in the field of named entity allocation in resumes presents several promising directions. Firstly, expanding the training dataset by incorporating a greater variety of resumes and domain-specific data could significantly enhance the accuracy of the algorithm. Additionally, exploring advanced concept mining and association mining techniques, tailored specifically to resume analysis, may improve the effectiveness of these approaches. Evaluating the algorithm on larger and more diverse datasets would provide a more comprehensive understanding of its capabilities. Furthermore, gathering user feedback and iteratively refining the algorithm based on their insights can lead to a more practical and effective tool for named entity allocation in resumes. By pursuing these research directions, we can advance the field and contribute to the development of more accurate and efficient algorithms for supporting HR processes and optimizing recruitment workflows.

% \section*{Acknowledgment}
% We would like to extend our heartfelt appreciation to Professor Dr. Y. V. Haribhakta for their unwavering support, guidance, and mentorship throughout this research project. Their expertise, valuable insights, and continuous encouragement have been instrumental in shaping the trajectory and outcomes of our work. We are truly grateful for their dedicated commitment to our intellectual growth and their invaluable contributions to our research. 

\bibliography{references}

\end{document}